\documentclass[10pt,twocolumn,letterpaper]{article}

\usepackage{cvpr}              % To produce the CAMERA-READY version
\usepackage{graphicx}
\usepackage{amsmath}
\usepackage{amssymb}
\usepackage{booktabs}

\usepackage[pagebackref,breaklinks,colorlinks]{hyperref}

\usepackage[capitalize]{cleveref}
\crefname{section}{Sec.}{Secs.}
\Crefname{section}{Section}{Sections}
\Crefname{table}{Table}{Tables}
\crefname{table}{Tab.}{Tabs.}

\def\cvprPaperID{587} % *** Enter the CVPR Paper ID here
\def\confName{CVPR}
\def\confYear{2022}

\begin{document}

%%%%%%%%% TITLE - PLEASE UPDATE
\title{Bending Graphs: Hierarchical Shape Matching using Gated Optimal Transport}

\author{Mahdi Saleh\textsuperscript{1}\thanks{the authors contributed equally to this paper}
\hspace{1.0cm}
Shun-Cheng Wu\textsuperscript{1}\footnotemark[1]
\hspace{1.0cm}
Luca Cosmo\textsuperscript{2,3} \\
\hspace{1.0cm}
Nassir Navab\textsuperscript{1}
\hspace{1.0cm}
Benjamin Busam\textsuperscript{1}
\hspace{1.0cm}
Federico Tombari\textsuperscript{1,4}\vspace{0.5cm}\\  
\textsuperscript{1}Technische Universit\"{a}t M\"{u}nchen \hspace{1.5cm} \textsuperscript{2} Ca’ Foscari University of Venice, Italy\hspace{1.5cm}\\  \textsuperscript{3}
USI University of Lugano, Switzerland \hspace{1.5cm}  \textsuperscript{4}Google\\[0.1cm] 
}

\maketitle

%%%%%%%%% ABSTRACT
\begin{abstract}
Shape matching has been a long-studied problem for the computer graphics and vision community. The objective is to predict a dense correspondence between meshes that have a certain degree of deformation. Existing methods either consider the local description of sampled points or discover correspondences based on global shape information. In this work, we investigate a hierarchical learning design, to which we incorporate local patch-level information and global shape-level structures. This flexible representation enables correspondence prediction and provides rich features for the matching stage. Finally, we propose a novel optimal transport solver by recurrently updating features on non-confident nodes to learn globally consistent correspondences between the shapes. Our results on publicly available datasets suggest robust performance in presence of severe deformations without the need of extensive training or refinement.
\end{abstract}

%%%%%%%%% BODY TEXT
\section{Introduction}
Deformable surfaces have been studied extensively by both computer graphics and computer vision communities. Dense correspondence estimation is closely linked to applications such as reconstruction and human pose estimation. In a standard formulation, we want to find a function $f:G \to H$ where $f$ is a mapping from a shape $G$ to another shape $H$. A shape is usually discretized in a triangle mesh that can be expressed in the form of a graph $G=(V, E)$ constituted by vertices $V$ with associated 3D coordinates and edges $E$.% 
Among the most successful optimization methods to find correspondences between deformable shapes, there is the Functional Map framework~\cite{functional}. The pointwise correspondence between two shapes is expressed as a linear map between the functional bases (i.e. eigenfunctions of the Laplace Beltrami Operator) defined on the shapes.%

\begin{figure}
\centering
\includegraphics[width=\linewidth]{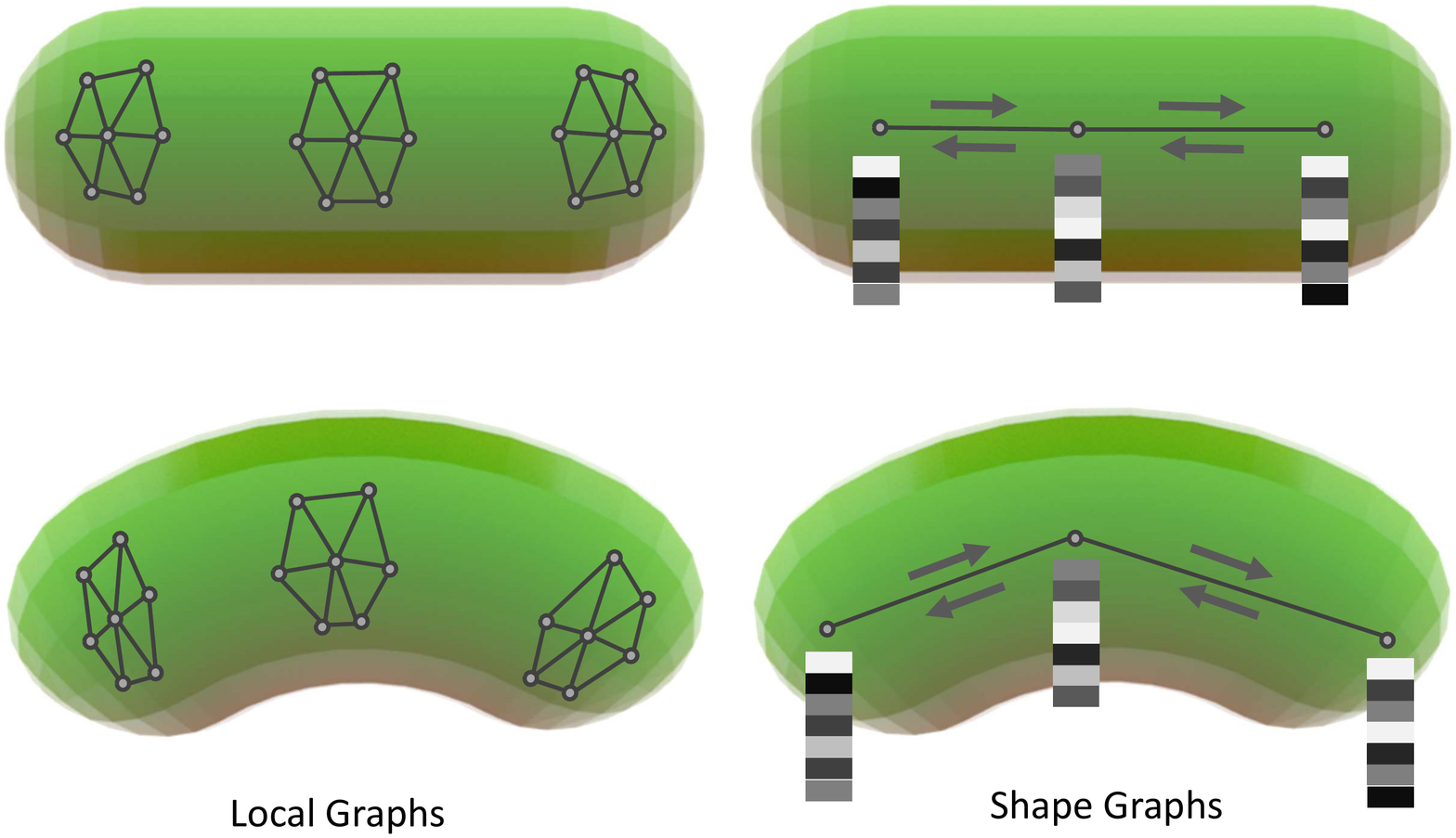}
\caption{We process deformable shape pairs by two levels of hierarchy, local graphs and shape graphs. Such hierarchical structure offers a flexible and holistic shape representation that enables correspondence matching and provides rich features for an optimal transport matching stage. }
\label{fig:teaser}
\end{figure}

As a first work to pair the functional map framework with deep neural networks on deformable shapes, deep functional maps~\cite{dfm} present a pipeline to seek a correspondence using as input SHOT~\cite{shot} handcrafted descriptors on sampled points. Despite building on top of point cloud features, FMNet~\cite{dfm} finds correspondences densely on deformable shapes. Recent works~\cite{3dcoded,trappolini2021shape} also use learned point cloud features~\cite{thomas2019kpconv} to describe local regions. Other methods also suggest integrating spectral manifold wavelets~\cite{hu2021efficient} or iterative spectral upsampling into functional maps~\cite{zoomout}. While these methods investigate dense correspondences, many require numerous initial keypoints or are sensitive to initial sparse matches.

Though meshes are commonly represented and stored as undirected lattice graphs, limited works explore graph deep learning frameworks to extract features from meshes~\cite{verma2018feastnet, monti2017geometric, gong2019spiralnet++}. Litany et al. use graph convolutional auto-encoders for the task of shape completion and Zhou et al. use graphs for image-based deformable matching. Graph neural networks (GNNs) are successfully applied to point clouds~\cite{wang2019dynamic,li2018pointcnn,te2018rgcnn,graphite} across many tasks. This work presents a graph-based descriptor designed for meshes, capturing local surface structures by constructing edges given a mesh lattice graph. GNNs on meshes are a discretization of spectral convolutions on manifold representations~\cite{bronstein2021geometric} and therefore constitute a powerful tool to capture local deformations.

Moreover, GNNs have proven to be a great framework to construct hierarchical representations \cite{wald2020learning,wu2021scenegraphfusion,cangea2018towards,ying2018hierarchical,farshad2021migs}. During recent years GNNs have been used in many computer vision tasks, from scene graph generation \cite{farshad2021migs} to holistic representation learning~\cite{zhou2021distilling}, enabling relationship definition \cite{kulkarni20193d} and information exchange and propagation~\cite{wu2021scenegraphfusion,xu2017scene}. Here we incorporate a high-level shape graph to represent and interconnect shape regions, as shown in figure~\ref{fig:teaser}. Each shape region describes the surrounding geometry by the local graph. Using such hierarchical graphs, we can process 3D meshes efficiently and learn a rich holistic shape representation, which is able to capture local details and capture neighbor geometry information.

We address the problem of 3D deformable shape matching as an optimal transport problem. As a classic problem, optimal transport is gaining popularity in the fields of feature matching and registration~\cite{sarlin2020superglue,yu2021cofinet,papakis2020gcnnmatch,puy2020flot}, where differentiable Sinkhorn algorithm~\cite{pai2021fast,sarlin2020superglue,yu2021cofinet} performs well with learning-based feature matching~\cite{chizat2020faster}. The Sinkhorn solver favors putative correspondences by iteratively applying softmax and does not work well in soft correspondence problems or when an exact matching is missing. The coarse deformable setting with sampled graph seeds does not guarantee putative hard matches. To deal with such an issue, we propose a new strategy using Gated Recurrent Units (GRUs) to propagate features using matching confidence in our shape graph leading to a more robust optimal transport solution.

In summary, our contributions are as follows:

\begin{itemize}
    \item We present a new representation for deformable meshes using hierarchical graphs, constituting of local feature graph and global shape graphs.
    \item We combine self-supervised local description and global shape features into an end-to-end deformable matching pipeline.
    \item We propose a novel Gated Optimal Transport (GOT) module incorporating attention-based feature propagation into the Sinkhorn algorithm.
\end{itemize}
\section{Related Work}
In this section, we will briefly review the related works in the area of 3D feature description, shape registration, and dense correspondence methods which are either related to the proposed method or are compared in the evaluation section.%

% \paragraph{Correspondence prediction}
\paragraph{Point descriptors}
Being able to produce a robust and descriptive point descriptor is at the core of many 3D Computer Vision tasks, especially when corresponding points have to be found between different objects. 
%3D descriptors are receiving increased attention thanks to the advances of RGB-D sensors. 
Unlike 2D image features, 3D features need to handle additional ambiguity introduced in the third dimension. To handle this ambiguity, some methods depend on a local reference frame, such as SHOT\cite{shot}, RoPS~\cite{guo2013rotational} and TriSi~\cite{guo2013trisi}, some others rely on pair-wise point description, such as PFH~\cite{rusu2008aligning} and FPFH~\cite{fpfh}. For non rigid meshes, spectral descriptors are often used \cite{bronstein2010scale,aubry2011WKS,cosmo2020average} given their invariance to \mbox{(near-)}isometric deformations. Later, data-driven approaches are proposed to compress hand-crafted features into a compact yet informative representation~\cite{khoury2017learning} or to learn a more robust feature description directly from point clouds~\cite{pointnet, ppfnet}. PointNet~\cite{pointnet} is the first approach that directly outputs feature description using points with a permutation invariant pooling, but it fails to capture geometric details and ignores neighbor information. PointNet++~\cite{qi2017pointnet++} is proposed to solve these issues by using multiple PointNets hierarchically to capture local details. 3DMatch~\cite{zeng20173dmatch} and Perfect Match~\cite{gojcic2019perfect} use voxel representation to compute feature descriptors with a 3D convolutional neural network, which is able to grasp the connection between voxels. The usage of 3D convolutions, however, significantly increases memory consumption and therefore limits the usability of these methods. Another line of works \cite{graphite,bogo2017dynamic,thomas2019kpconv} benefits from the point representation while using GNNs to incorporate the information from surface and manifolds. 
KPConv~\cite{thomas2019kpconv} defines 3D local filters with a set of kernel points which allow efficient and flexible point description. Graphite~\cite{graphite} uses a Graph Neural Network to describe local patches and predict keypoints for point cloud registration.%

\paragraph{Shape registration}
Shape registration aims to find the deformation between two given geometric shapes. A classic solution is ICP~\cite{icp} which estimates a transformation matrix to minimize the distance between closest points of the given shapes iteratively. Several methods are proposed to improve the matching performance of ICP, such as GoICP~\cite{goicp}. PointNetLK~\cite{pointnetlk} exploits learned point features~\cite{pointnet} with the Lucas-Kanade algorithm~\cite{lk} to tackle the registration problem. Deep Closest Point~\cite{dcp} incorporates the popular attention mechanism in the correspondence finding process to estimate the transformation. 
Other approaches reinforce correspondence prediction leveraging additional sources of knowledge, such as parametric human~\cite{loper2015smpl,pishchulin2017building, pons2015dyna,faust} or animal~\cite{zuffi20173d, zuffi2018lions,zuffi2019three} models.%

\paragraph{Dense correspondence}
In the correspondence problem, the goal is to find a map relating the points of an input shape to the points of a second one, possibly undergoing some deformation. Most of the methods for non-rigid shape matching ~\cite{kim2011blended, functional,rodola2017partial} relies on intrinsic properties of the input shapes.
%~\cite{kim2011blended, functional} estimate dense correspondences between two given shapes by realizing the intrinsic properties of the given 3D surfaces. 
%These methods explicitly or implicitly assume an isometric deformation of the inputs. 
A common drawback of these methods is that correspondences are poorly localized. Moreover, intrinsic properties are unaware of isometrics in the input shapes, resulting in misplaced matches.
%A common drawback of these methods is that the initial descriptor, which computes the eigenfunctions of the Laplace-Beltrami operator, is usually unstable and difficult to estimate.
Several methods have been proposed to solve these issues~\cite{vestner2017product,gasparetto2017spatial,zoomout}, but still, the performance deteriorates in a more challenging scenario~\cite{shrec19}. Unlike previous works that rely on axiomatic descriptors~\cite{rodola2017partial,nogneng2018improved}, recent methods focus on learning an optimal descriptor to have a better functional map estimation~\cite{dfm,cosmo2016matching, halimi2019unsupervised,dgfm,roufosse2019unsupervised, sharp2020diffusion, trappolini2021shape, marin2020correspondence}. Marin et al.~\cite{marin2020correspondence} propose a two-stage method to estimate an optimal linear transformation by the use of an invariant embedding network combined with a probe function network. Trappolini et al.~\cite{trappolini2021shape} propose to estimate the transformation between two input point clouds with an auto-encoder architecture and a transformer network~\cite{transformer}. Groueix et al.~\cite{3dcoded} learn to estimate the transformation between two given shapes using a neural network. 
\section{Bending Graph Methodology}
\begin{figure*}
    \centering
    \includegraphics[width= \linewidth]{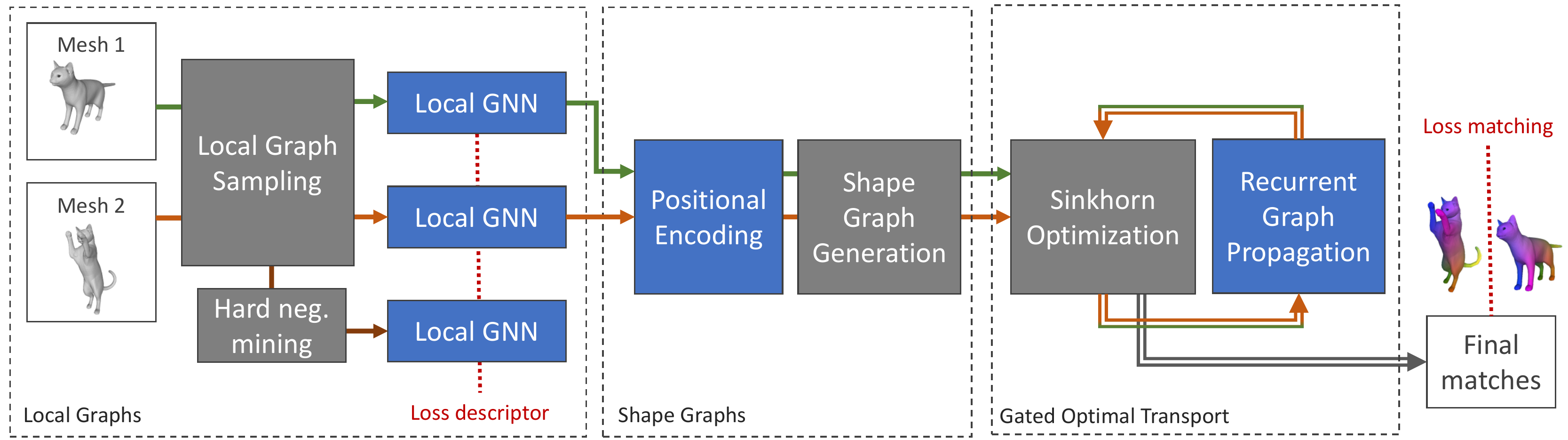}
    \caption{Our pipeline consists of 3 stages: 1) (left) Local graphs are sampled over a mesh pair. A negative sample is used for training the descriptor. 2) (middle) the shape graphs are constructed, consisting of local GNNs with spatially connected edges; the positions are also encoded into the shape graphs. 3) (right) a GOT learner integrates a Sinkhorn optimizer into a recurrent graph propagation unit and produces confident matches.}
    \label{fig:overview}
\end{figure*}
In this section, we illustrate our methodology. The overview of our method can be seen in figure~\ref{fig:overview}. As a standard dense matching pipeline between two shape meshes, we first focus on each input mesh and process them independently. We start by explaining our local graph definition and how the descriptor is configured in section~\ref{local}. Next, in~\ref{shape}, we discuss how the higher-level shape graph is constructed to represent the shape structure. In section~\ref{GOT}, we focus on the matching problem where we present a GOT unit to predict the correspondences. At the end of this section in \ref{losses} we explain the training and loss formulations.%
\subsection{Local Graphs}\label{local}
Each of the input shapes is given in the form of a mesh. A mesh $M$ consists of vertices $V \in \mathbb{R}^3$, edges $E$, and $M=(V,E)$ which form a regular tiling. This definition is very close to that of an undirected graph, where a node represents vertices. To define local structures, we focus on sampled local graphs $G_i, i \in \mathbb{N}$. To sample $N$ graphs, we first apply furthest point sampling (FPS) on mesh vertices in Euclidean space. We form graphs $G_i$ around each of the sampled vertices $v_j \in V$. In order to construct the graph based on the mesh structure, we define edges using the local Dijkstra algorithm. We cut a sub-graph from a mesh graph where the vertices are below a certain shortest path value $d_{cut}$:
\begin{equation}
    V_i=\{v_j \in G \mid d(v_i,v_j)<d_{cut}\},
\end{equation}
where $d(v_i,v_j)$ defines the shortest paths between the vertices $v_i$ and $v_j$. Each node is associated by its coordinate values in a local reference frame, $v_j=(x_j,y_j,z_j)$.
Edges are linked to nearby vertices. As an alternative to KNN~\cite{bogo2017dynamic}, we consider the unit ball with radius $r$ to maintain metrics around each node's positional coordinates and connect them to other nodes when their distance falls bellow $r$. To induce some weight to the edges, we attribute a scalar value $e_{i,j} \in \mathbb{R}$ to it as the vertex-to-vertex distance.%

\paragraph{Local Mesh Description} We now introduce our graph neural network (GNN) architecture. We build graphs consisting of edges and nodes to represent a local mesh. In contrast to classic and learned point cloud descriptors, which only define a certain number of points~\cite{graphite,shot,ppfnet}, here we want to describe dynamically-sized graphs based on how the local mesh is structured.%

The node and edge features,  connections are fed into a GNN architecture to estimate a descriptor under variable deformations. Output descriptors should be invariant to permutations, meshing variations, and applied deformations. Following Graphite~\cite{graphite} we use a topology adaptive graph (TAG)~\cite{du2017topology} convolutional operator, which combines node and edge feature propagation inside the graph. We use multiple layers of TAG function with increasing hops (K=1,2,3). Hops define how many nodes the information propagates inside the graph and provides a multi-scale feature extraction by message passing. Every graph convolutional module takes into account the adjacency matrix $\mathcal{A} \in {\rm \mathbb{R}}^{n \times n}$ and its diagonal degree matrix $\mathcal{D} \in {\rm \mathbb{R}}^{n \times n} $ to propagate node features across the graph. As in~\cite{du2017topology}, we update node-level information $v_j'$ by propagating features as follows:
\begin{equation}
v_j^{\prime} = \sum_{k=0}^K \mathcal{D}^{-1/2} \mathbf{\mathcal{A}}^k
        \mathcal{D}^{-1/2}v_j \mathbf{\Theta}_{k}.
\label{eq:tag}
\end{equation}

Following the multi-scale message passing in each local graph, we apply a global max pooling operator to extract the local feature descriptor $D_i$. The features are then passed to a Multi-Layer Perception (MLP) and normalized to bring the final feature on a unit sphere. The local features $D_i, i \in \!N$ are trained as a triplet to encourage more robustness of features as further explained in \ref{losses}.%

\subsection{Shape Graphs}\label{shape}
In this section, we describe our higher-level shape graph. The intention of the shape graph is to build a coarse-level representation of the shape and facilitate feature propagation between the local graphs. In this way, we can incorporate more global features into the node descriptor and provide better grounds for correspondence matching.%

A shape graph $S$ consists of $N$ nodes, where each node is associated with a local descriptor vector $D_i \in \mathbb{R}^d$ and seed point $ v_i \in \mathbb{R}^3$, initially sampled in \ref{local} using FPS. Furthermore, we define edges using a unit ball in the geodesic space. Keeping the shape into a unit ball, we further define a shape radius $r_{shape}$ to construct edges between shape nodes. Similarly to local graphs, we add an edge weight based on the geodesic distance of the nodes. Figure \ref{fig:shapegraphs} shows sample shape graphs generated from the MPI FAUST~\cite{faust} dataset. The graph structure remains somewhat similar in presence of local deformations. %

\paragraph{Positional Encoding} In order to aggregate the node information, we need to present the position in a high-dimensional embedding. This process is commonly performed using layers of MLP similar to~\cite{pointnet}. The absolute input position $v_i$ should be encoded to a feature of size $d$. To capture the fine details and inspired by NeRF positional encoding~\cite{mildenhall2020nerf}, we use a Fourier feature mapping~\cite{tancik2020fourier}. We map input coordinates into a higher dimensional Fourier space before passing them through the network with
\begin{equation}
    \gamma(v_i)=\left[ \ldots,\cos(2\pi \sigma ^{j/m} v_i),\sin(2\pi \sigma^{j/m} v_i),\ldots \right]^T.
\end{equation}

We extract $m-1$ log-linear spaced frequencies for each positional element. Afterwards, we pass them to a shallow MLP network ($\text{MLP}_e$) to create an embedding of size $d$. The embedded positional encoding is added to $D_i$, followed by another MLP network ($\text{MLP}_p$) to obtain the node feature $f_i$. The entire process can be expressed as follows:
\begin{equation}
    f_i = \text{MLP}_p \left( D_i + \text{MLP}_{e} \left(\gamma(v_i) \right) \right).
\end{equation}
\subsection{Gated Optimal Transport}\label{GOT}To find the matching between two 3D meshes, we formulate the task as a linear assignment problem, which tries to maximize the total score of $\sum_{i,k} C_{i,k}P_{i,k}$ with an assignment $P$ and a score matrix $C\in \mathbb{R}^{M\times N}$. As in~\cite{sarlin2020superglue}, the score for each match is calculated as a simple inner product of their descriptors:
\begin{equation}
    C_{i,k} = \left \langle f_{i}^{A}, f_{k}^{B}  \right \rangle, \forall(i,k)\in A \times B,
    \label{eq:op_score}
\end{equation}
where $\left \langle \cdot,\cdot\right\rangle$ is the inner product, $A$ and $B$ represent a source and a target.

The above optimization problem can be efficiently solved with the Sinkhorn algorithm. This algorithm estimates bipartite joint probabilities by iteratively normalizing $\exp(C)$ along rows and columns with a given number of iterations. Since the entire operation is differentiable, the whole process can be trained end-to-end by minimizing the negative log-likelihood of $P$. The loss formulation will be explained in \ref{losses}.

\paragraph{Gated Feature Propagation}
\begin{figure}[h]
\centering
\includegraphics[width=0.8 \linewidth]{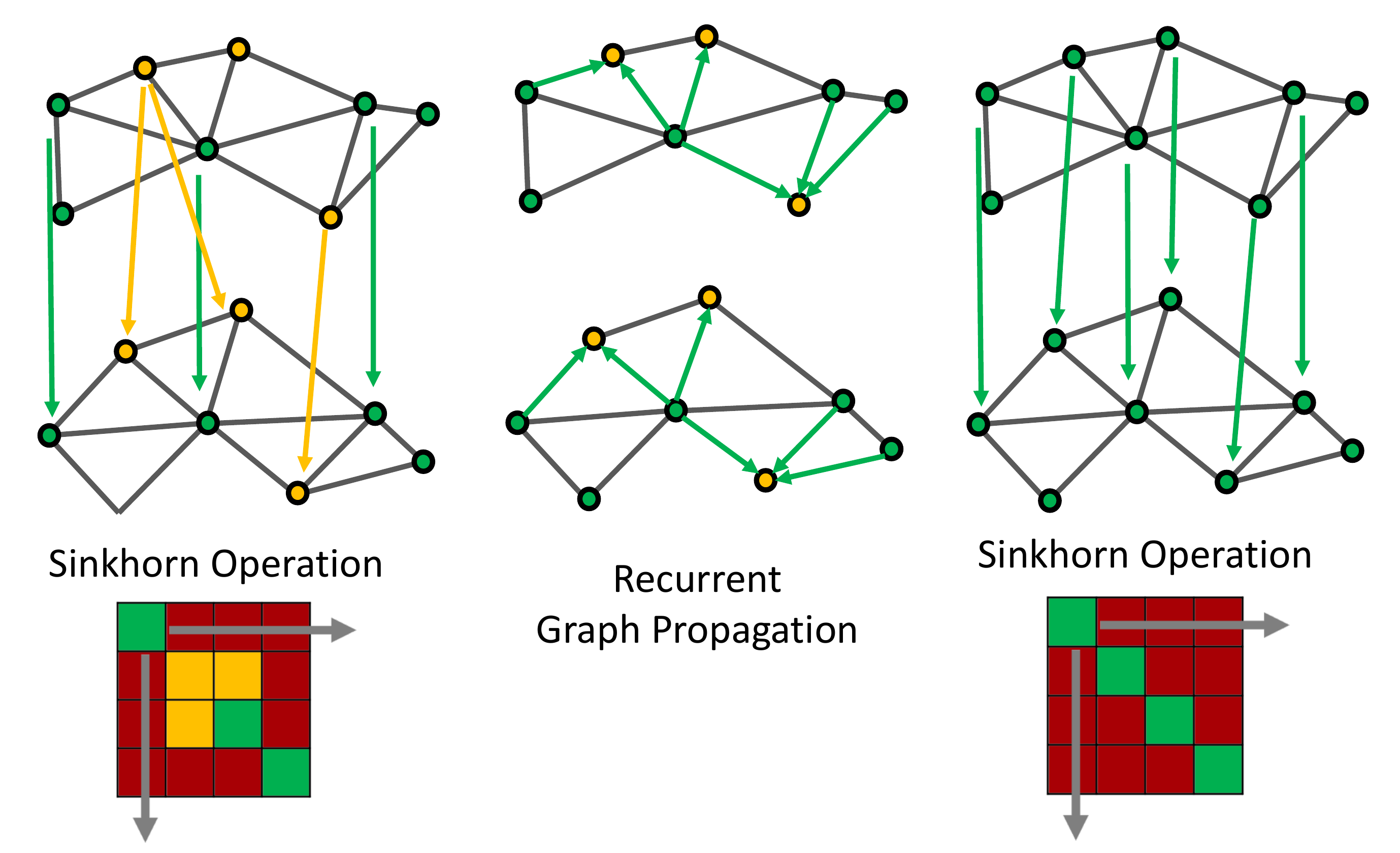}
\caption{Our GOT module consists of Sinkhorn iteration and recurrent feature propagation (in middle) in the shape graph. Promoting the feature propagation of more confident nodes, lets more information flow into weak nodes, consequently improving graph matching.}
\label{fig:got}
\end{figure}

% mention a bit on what is the issue
Although solving the optimal transport with Sinkhorn provides a fast and direct way to solve the bipartite matching, it focuses on finding the exact match on each given input without considering the mesh topology. Two nearby vertices on a mesh may be assigned to different regions after the matching process. To tackle this issue, we propose to utilize the confidence in the Sinkhorn operation to effectively propagate the feature with high confidence to its nearby low confidence features along with the shape graph, which we call 
GOT. This enforces the nearby point features to be matched to a nearby locations on the target region, as shown in Fig.~\ref{fig:got}.

Node feature propagation on the shape graph with the confidence value estimated by the similarity score can be considered as conditional random fields (CRFs)~\cite{lafferty2001conditional}, which is typically the final optimization step in a pipeline. Previous work showed that similar processes are achievable at training time using recurrent neural networks (RNNs) \cite{zheng2015conditional, xu2017scene}. Inspired by these, we formulate the node feature as a mean-field approximation with the connectivity as the shape graph. Similarly, we use mean-field to perform approximate inference. The feature of a node on each update step can then be formulated as:
\begin{equation}
    h_i^t = Q\left(h_{i}^{t-1},  \max_{l\in \mathcal{N}(i)} \left( w_l h_{j}^{t-1} \right)  \right),
\end{equation}
where $h^{t}_i$ is the hidden state of node $i$ at time $t$, $\mathcal{N}(i)$ is all the neighbors of node $i$,  $Q$ is an RNN model, for which we used a GRU~\cite{cho2014properties} unit and $w_l$ is the weight of node $l$. The initial hidden state of each node $h_{i}^{0}$ is initialized as $f_i$.

As each node may receive multiple features from its neighbors, we weight each input by the log likelihood estimated in the score matrix $C$. The weight value of each node is estimated in the Sinkhorn assignment, which can be obtained from the score matrix $C$ by:
\begin{equation}
    w_i = \max_l\left( C_{i,l} \right),\qquad w_l = \max_i\left( C_{i,l} \right).
\end{equation}
The confidence values are kept in log space in the message passing process, since we found that it results in better performance. 
The usage of the shape graph allows us to control the number of hop neighbors to be considered in the message passing operation. Furthermore, the entire GOT process can also be applied iteratively to reinforce the optimal transport solution.%

\subsection{Loss Functions}\label{losses}
In this section, we describe our training strategy and loss functions. As expressed initially in section \ref{local}, we are given pairs of shapes as meshes, and we suggest a mapping for each vertex in $V_a$ to a vertex in $V_b$. To enable efficient and coarse-level matching, we sample $N$ points from each shape. To train our descriptor using a triplet loss, we also require a negative sample. As a common practice in metric learning and to increase feature distinctiveness, we add a hard-negative sample from shape $B$. To mine a negative sample, we use the Dijkstra algorithm to draw a graph in the vicinity of the target seed vertex.%

\paragraph{Local descriptor}
Learned descriptors or features are trained by self-supervision of pose or deformation variations~\cite{ppfnet,graphite,3dcoded}. The estimated feature vector can be trained using contrastive or metric learning. We make sure a feature pair $D(G_a)$ and $D(G_b)$, describing respectively the local graphs of $G_a$ and $G_b$, are closed in a feature space. In a triplet setting, we furthermore use a negative $D(G_n)$ describing a negative graph sample as explained below: 
\begin{equation}
    \mathcal{L}_D = \max\left(\sum_{i=1}^{N} \text{dist}(G_{a,i},G_{b,i})- \text{dist}(G_{a,i},G_{n,i}) + \alpha,0\right)
\end{equation}
where $\text{dist}(G_{a,i},G_{b,i}) = ||D(G_{a,i})-D(G_{b,i})||_{2}$
and $\alpha$ defines a small margin to reduce zero values in the loss.%

\paragraph{Matching} Following our GOT module, described in \ref{GOT}, we yield a cost matrix of size $N \times N$, where each index $C_{i,l}$ defines the softmax confidence value derived from Sinkhorn iterations. During training, we have a bipartite distance matrix $\mathcal{M}$, where each $\mathcal{M}_{i,l}$ defines the shortest path in the graph from node $i$ from shape $A$ to node $l$ from shape $B$. Close matches are defined as minimum entries in the bipartite distance matrix. Our final goal is to have maximum score values on such entries.%

In order to increase the log-likelihood of the entries softly based on the bipartite distance matrix, we define a weighting matrix, $M^{\prime}_{d}$ based on the matrix $\mathcal{M}$ where
\begin{equation}
    \mathcal{M}^{\prime}_{i,l} =
    \begin{cases}
    \frac{r_d - \mathcal{M}_{i,l}}{r_d} ,& \text{if } \mathcal{M}_{i,l}\leq r_d\\
    0,              & \text{otherwise}
\end{cases}
\end{equation}
where $\mathcal{M}^{\prime}_{i,l}$ defines an element of matrix $\mathcal{M}^{\prime}_{d}$. We use this weight to induce some distance softly into the matching loss. Our loss is minimising the negative log-likelihood of correct matches on the nonzero elements of $\mathcal{M}^{\prime}$. 
\begin{equation}
    \mathcal{L}_m = - \sum \log \left( v_{i,l} \cdot \mathcal{M}^{\prime}_{i,l} \right)
\end{equation}%

\paragraph{Regularization term} The predicted match for node $i$ of A in B is where $max_{l}v_{i,l}$. This is how we define our matching loss in the previous section. However, to regularize the predictions further and use the global shape structure to assert shape consistencies, we can use conventional graph operators to assert local gradient consistencies. We first apply a softpooling operator on predicted positions. This way, we can have a differentiable operation for the loss calculation. We first calculate
\begin{equation}
    \hat{s_i} =\frac{1}{N} \sum_{j}^{N}{C_{i,j}\cdot v_{j,j}}
\end{equation}%
and then define a Laplace operator $\Delta(V)_i$ on the original shape $V$ with source positions $v_i \in \mathbb{R}^3$ and predicted softpool positions $\hat{V}$ as 
\begin{equation}
    \Delta (V)_i = \sum_{J:(i,j)\in E}|v_j -v_j|,
\end{equation}
where $E$ is the set of shape graph edges.

Finally, we propose our regularization loss as follows:
\begin{equation}
    \mathcal{L}_R = \sum_{i=1}^{N}{\Delta(\hat{V})_i -\Delta(V)_i }
\end{equation}
Our total loss is a summation of the discussed loss functions:
\begin{equation}
    \mathcal{L}_{total} = \gamma _D \cdot \mathcal{L}_D +  \gamma _M\cdot \mathcal{L}_M + \gamma _R \cdot \mathcal{L}_R
\end{equation}
\section{Experiments}
\subsection{Training Setup}
In this section, we demonstrate the performance of our method by showing two major evaluations on popular datasets used by the state-of-the-art methods for deformable 3D shape correspondences. The first experiment is to evaluation our network on the task of human shape registration (sec.~\ref{sec.exp1}), and the second one is on the task of animal shape registration (sec.~\ref{sec.exp2}).
We further ablate each of the proposed modules both qualitatively and quantitatively (sec.~\ref{sec.abla}). 

\begin{figure}
    \centering
    \includegraphics[width=1.0 \linewidth]{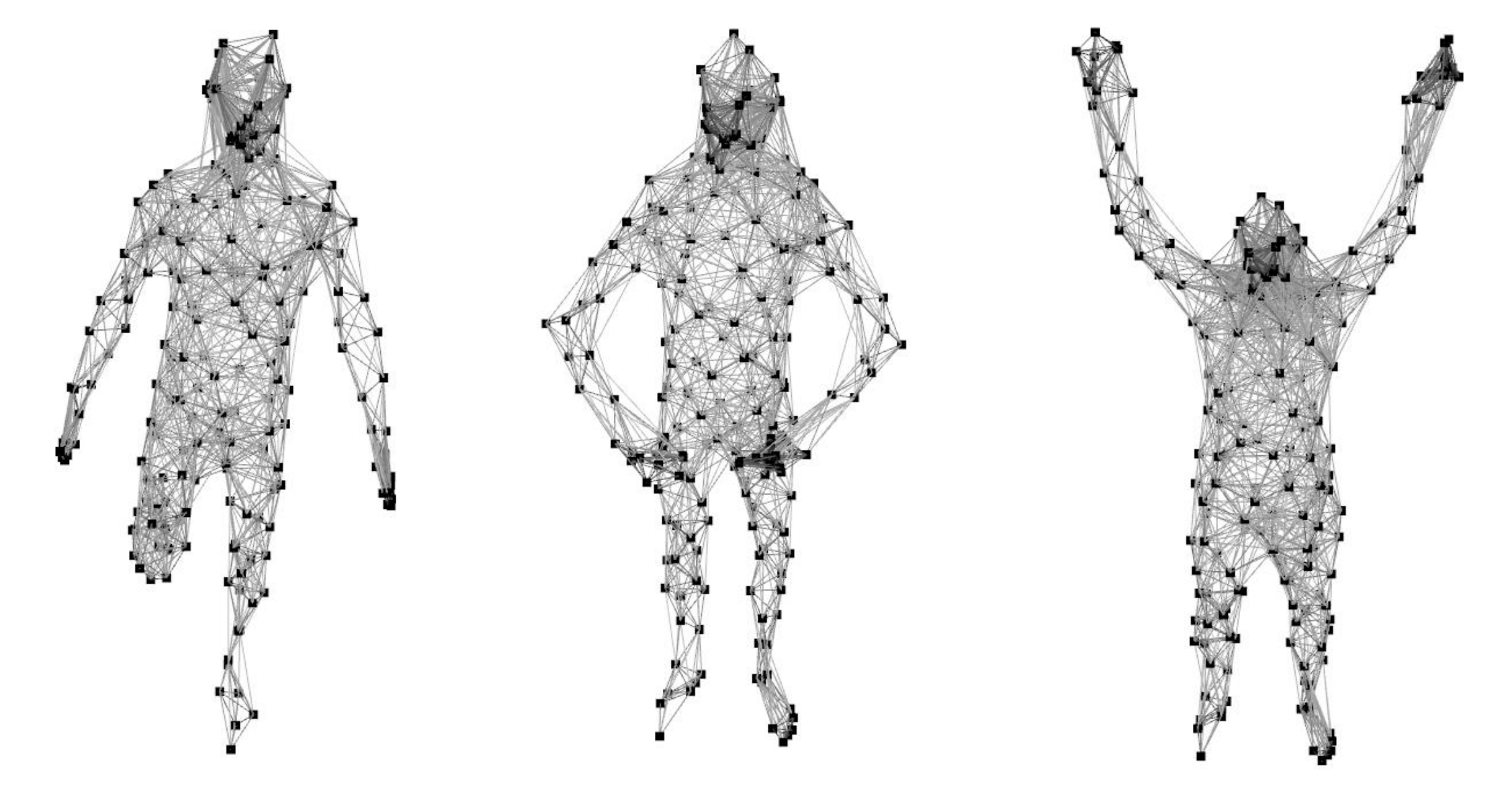}
    \caption{Samples of Shape Graphs from MPI-FAUST \cite{faust} dataset. We sample 200 local graphs and associate local graphs spatially using the depicted shape graphs. The graph edges are weighted based on the pairwise Euclidean distance (visualized with grayscale edge color). Under severe deformations, the graph shape still shows consistent structure. }
    \label{fig:shapegraphs}
\end{figure}

\subsection{Experimental Settings}
In the first experiment, we use the FAUST dataset~\cite{faust} with the same testing split used in~\cite{marin2020correspondence, trappolini2021shape}. This dataset contains 100 human shapes with a per-vertex correspondence which allows us to evaluate the dense correspondence quality of our method. For training, we generate 500 samples using the SURREAL dataset~\cite{surreal}, which consists of human SMPL~\cite{loper2015smpl} models with a set of parameters to control the deformation and poses of the models. 
In the second experiment, we use the animal shapes provided in TOSCA~\cite{tosca}. The TOSCA dataset provides several synthetic models in different poses and classes. For testing, we consider all pairs composed of the T pose of each class with all other poses in the same class. For training, we generate 100 random models from SMAL~\cite{zuffi20173d} with a Gaussian distribution of variance 0.15. 

In all experiments, we sample 200 local graphs with farthest point sampling (FPS), as shown in figure~\ref{fig:shapegraphs}. Note that the sampled points on the source and target mesh may not be in the same location. Therefore we do not have exact and putative correspondences during inference. Therefore, dense correspondence matching is achieved by using the functional map method~\cite{functional}. The error metric in all experiments follows~\cite{kim2011blended} which uses the average geodesic error. For ablation study, we further report the Bijectivity Rate (BR) measures the percentage of bijective correspondences between source and target (one-to-one consistency) over the total number of correspondences.

For all the experiments, we implemented our method on Pytorch and trained with an initial learning rate of 0.001, decreasing to 0.0001 after 30 epochs with ADAM optimizer. %For experiment 1, we only generate 500 training samples.

\begin{figure*}
    \centering
    \includegraphics[width=\linewidth]{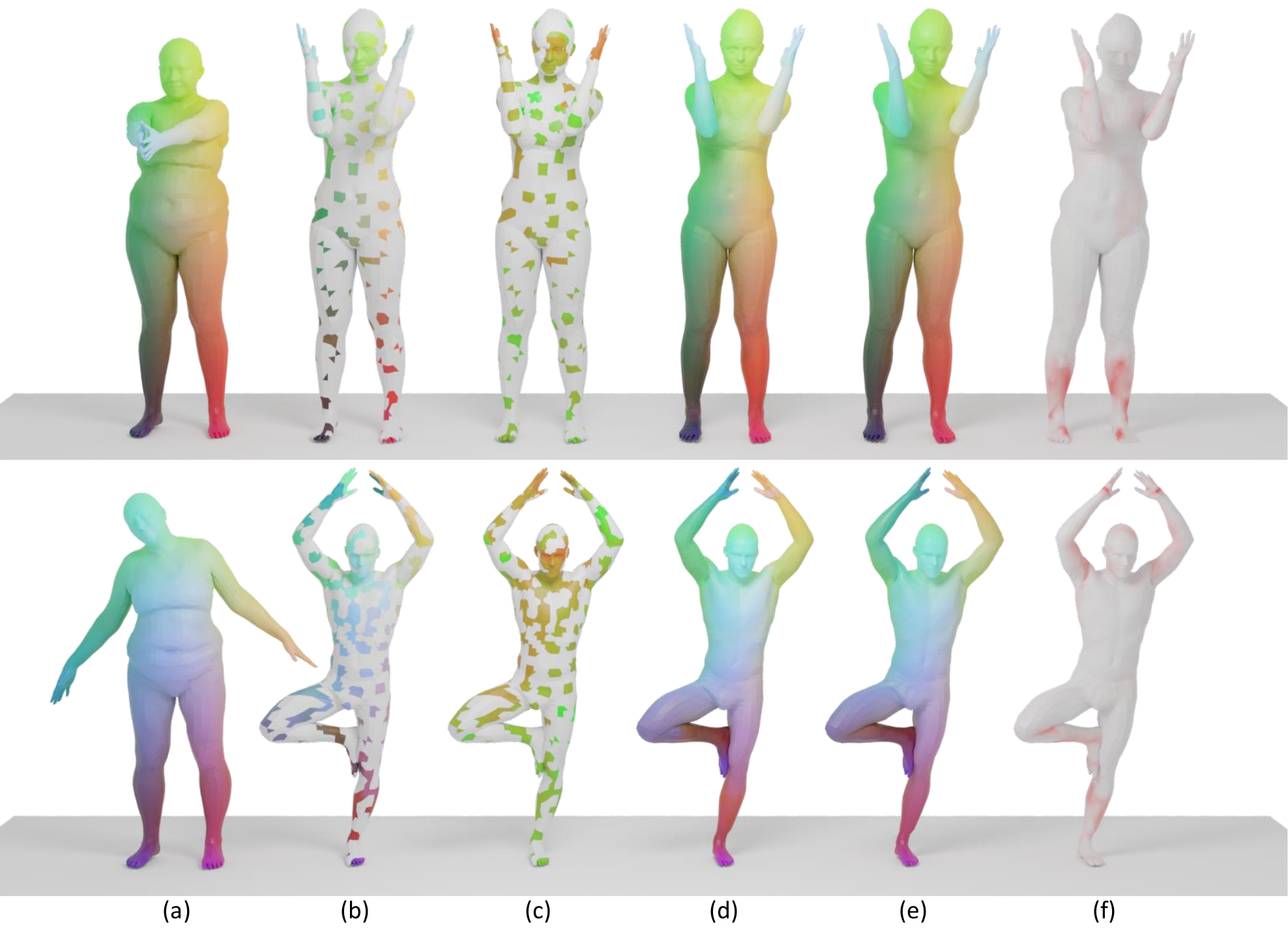}
    \caption{The result of dense shape matching on MPI-FAUST\cite{faust} trained on a few samples of SURREAL\cite{surreal}. (a) we have a target shape to which we are finding correspondences. Coarse matches (b) show robust matched patches to the target shape. In (c) we visualize the final confidence values predicted. The maps suggest more confidence in less deformed regions. In (d) we can see our fine and dense correspondences by passing our coarse matches to the FM algorithm in form of corresponding delta functions. Finally, we have the Ground correspondence map in (e) and error map in (f) which highlights our good performance for inter-subject cases without the need for refinement.}
    \label{fig:faust}
\end{figure*}

% In \cite{trappolini2021shape}, they trained the mode on the first 1k shapes from SURREAL~\cite{surreal}, and tested on the rest 2700 and on FAUST1K. 
% The ablation they did includes the number of latent space vectors, their dimension, and the number of encoder-decoder blocks.

\subsection{Human Shape Registration}\label{sec.exp1}
We compare our method with two dense correspondence methods using an auto-encoder architecture, \ie 3DC~\cite{3dcoded} and SRT~\cite{trappolini2021shape}, a functional map method with learned linear high dimensional basis, \ie LinInv~\cite{marin2020correspondence}, and a method that directly learns optimal descriptors with the functional map framework, \ie DGFM~\cite{dgfm}. The dense correspondence result from our method is generated by first estimating the coarse correspondences on the 200 patches sampled with FPS, and then giving this correspondences as input to the functional map algorithm in form of corresponding delta functions. The result is shown in table~\ref{tab:surreal_faust}. It can be seen that our method outperforms previous methods by a significant margin. We also report the number of parameters used in each method. Due to the use of a lightweight GNN module and our hierarchical design, our method needs only 100k trainable parameters, which is $0.7\%$ used in~\cite{dgfm} and $7.1\%$ of the parameter used in~\cite{marin2020correspondence}.
The qualitative result is shown in figure~\ref{fig:faust}. Our method can match patches correctly in locations that undergo significant deformations.

% In \cite{trappolini2021shape}, they trained their model on the Surreal dataset~\cite{surreal} used in~\cite{marin2020correspondence}. It consists of 10000 point clouds for training. Each point cloud has 1000 points. During training, they augment the data by randomly rotating shapes along the second axis. For testing, they use the 1K point version of FAUST dataset~\cite{faust} also from~\cite{marin2020correspondence} in which the points are perturbated by Gaussian noise. In addition, they also tested their model on SHREC'19~\cite{shrec19}, which consists of 44 shapes that have different connectivities, poses, and densities. 

\subsection{Animal Shape Registration}\label{sec.exp2}
\begin{figure}
    \centering
    \includegraphics[width=1.0\linewidth]{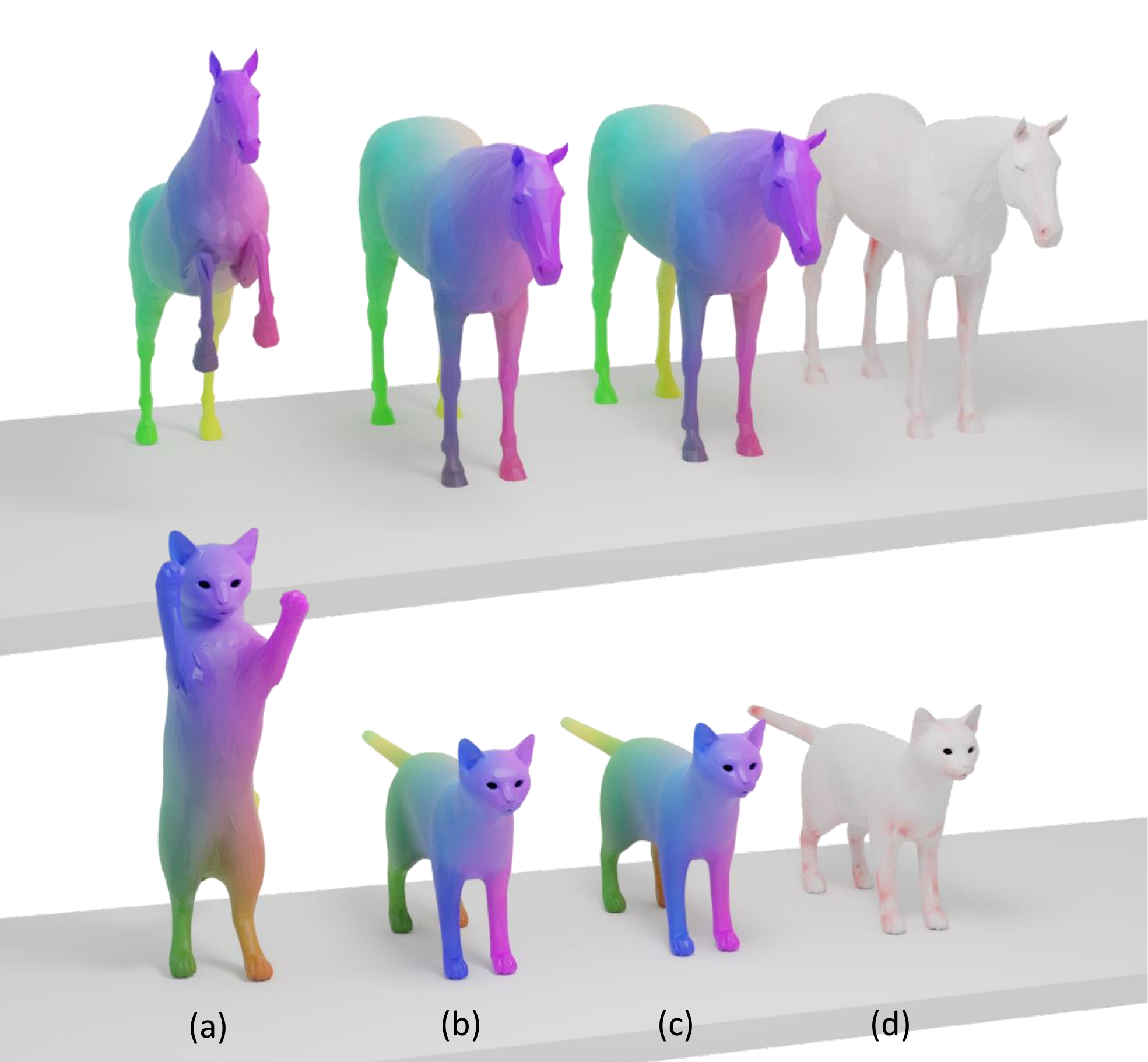}
    \caption{The dense matching result of our model on TOSCA dataset. Our method is able to learn robust features that work on an unseen class. a) target shape. b) source shape with the predicted correspondences c) ground-truth correspondences, d) error map.}
    \label{fig:tosca}
\end{figure}

To show our method can be used across different shapes, we evaluate our method on the TOSCA dataset. For this evaluation, we trained a model with 100 random samples on the horse class from SMAL and tested it on horse, cat, centaur and david from TOSCA. The results are shown in figure~\ref{fig:tosca} and table~\ref{tab:tosca}. The dense correspondence results suggest that our method can also be applied to other shapes. The matching on unseen classes is more challenging since the class was unseen during training. Thanks to our hierarchical design, the model can generalize to unseen shapes.

\begin{table}
    \centering
    \begin{tabular}{rcc}
    \toprule
         &  coarse error ($\downharpoonright$) & fine error ($\downharpoonright$) \\
     \midrule
     horse    & 0.2194 & 0.0231 \\
     cat      & 0.1856 & 0.0143 \\
     centaur  & 0.2150  & 0.0193 \\
     david &  0.3254 &  0.0213 \\
     \bottomrule
    \end{tabular}
    \caption{The average geodesic error on TOSCA dataset~\cite{tosca}. The coarse error is calculated after the patch matching of our method. The fine error is calculated by feeding the estimated correspondences to the functional map. Note that the model is trained only on the horse class and has never seen other classes.}
    \label{tab:tosca}
\end{table}

\setlength{\tabcolsep}{2pt}
\begin{table}
    \centering
    \begin{tabular}{lcc}
    \toprule
        Method &  geodesic error & parameters~(million)\\
    \midrule
        3DC~\cite{3dcoded} & 0.0776 & 3.9\\
        DGFM~\cite{dgfm} & 0.0656 & 14.1\\
        LinInv~\cite{marin2020correspondence} & 0.0942 & 1.4 \\
        SRT~\cite{trappolini2021shape} & 0.0513 & 1.7\\
        % Transformer~\cite{trappolini2021shape} w. r. & & 0.0369 \\
        Ours & \textbf{0.0230} & \textbf{0.1}\\ %0.02295
    \bottomrule
    \end{tabular}
    \caption{We trained all models on SURREAL \cite{surreal} models and tested them on MPI-FAUST synthetic~\cite{faust}. Contrary to previous methods, we only train our model on 5\% of the available dataset.}
    \label{tab:surreal_faust}
\end{table}%

\subsection{Ablation Studies}\label{sec.abla}
To prove the functionality and performance of each proposed module, we design ablation studies as a further experiment. Table \ref{tab:ablation} contains the comparative results of the modules by looking at bijectivity rate (\%) and average geodesic error normalized to the shape area. We follow the experiment setting in \ref{sec.exp1} by training our pipelines on 100 samples of SURREAL and testing on 20 samples of MPI-FAUST. Ablation \#1 shows matching with Vanilla OT and only using local description (LG) does not provide reliable correspondences. By adding the shape graph (SG) in ablation \#2, we observe a performance gain in terms of geodesic error. Here we use a Sinkhorn algorithm with 100 iterations. In this case, the correspondences contain a few bijective matches. Adding a GOT layers in ablation study \#3 yields more bijective correspondences, and improved final error. Note that the error is calculated on all the points without any outlier removal.%

Next, in ablation \#4 we deactivate the shape graph, which weakens our graph features. Nevertheless, compared to \#1, we see slightly better performance. This ablation proves the effectiveness of our fine Fourier-based positional encoding. Experiment \#5 illustrates the role of local graphs and descriptors and proves the local descriptor would be richer with global holistic knowledge. Ablation \#6 shows the full pipeline. By comparing \#3 and \#6, we can observe less error and higher bijectivity which proves the efficacy of the regularization loss.
\setlength{\tabcolsep}{2pt}
\begin{table}[!htbp]
% \scriptsize
    \centering
    \begin{tabular}{lccccrr}
    \toprule
         Experiment & LG &  SG & GOT & Reg & BR & error\\
    \midrule
        \#1 only local desc., w. OT &\checkmark & & & & 42.1 & 0.58 \\
        \#2 hier desc.,  w. OT &\checkmark &\checkmark & & & 23.3 & 0.42 \\
        \#3 w/o graph regularization &\checkmark &\checkmark & \checkmark& & 40.0 & 0.19 \\
        \#4 w/o shape graphs &\checkmark & & \checkmark& \checkmark &  21.1 & 0.43  \\
        \#5 w/o local desc. & &  \checkmark &  \checkmark& \checkmark & 47.0 & 0.16\\
        \#6 full pipeline &\checkmark &\checkmark &  \checkmark& \checkmark & 49.6 & 0.13\\
    \bottomrule
    \end{tabular}
    \caption{Ablation study trained on SURREAL~\cite{surreal} and evaluated on MPI-FAUST~\cite{faust}. LG stands for activation of Local Graph unit \ref{local}; SG for Shape Graph module \ref{shape}; GOT for our Recurrent Graph Propagation unit and Reg. for activation of our regularization loss. For each ablation, we provide the percentage of bijective correspondences (Bij. rate) and average geodesic error.  }
    \label{tab:ablation}
\end{table}

\section{Limitation and failure cases}
% maybe that we need the corresponding points during training, the source and target should have almost the same number of points, otherwise subsampling
% show the sample train on Surreal and test on SHREC'19. Just show the error map. 
To learn and estimate correspondences, we suppose a pair of mesh with similar density. When the number of vertices varies, we need to reconstruct the mesh as a preprocessing stage to describe and match the meshes independent of the number of the points. Moreover, in our training and experiments, we always normalize the meshes to the unit spheres. Although this is common practice in deformable registration, we do not provide a scale-invariant representation. The main reason for normalization is setting a fixed radius for ball query operations across different datasets. Finally we only learn the coarse matches using our representation and rely on further non-learning refinements to provide dense matches.
\section{Conclusion}
In this paper, we propose Bending Graph, an end-to-end pipeline to learn deformable shapes in a hierarchical form. Hierarchical graphs can represent the shape flexibly and efficiently. With the use of Local and Shape Graphs, we learn object representation in a holistic way that can integrate into a matching pipeline. Furthermore, we look at the problem of dense matching using Optimal Transport. We propose a solution to learning-based Optimal Transport using Gated Recurrent Network. Using our representation, we can propagate and reinforce our features through the shape graphs. Finally, we demonstrate our pipeline and prove its effective design by providing robust results without the need for large-scale training and computationally. Our representation and matching framework can be used for multiple problems in computer vision and graphics.
\newpage
{\small
\bibliographystyle{ieee_fullname}
\bibliography{ms}
}

\newpage
% \title{Supplemental Materials}
\textbf{\large Supplemental Materials}

\setcounter{equation}{0}
\setcounter{figure}{0}
\setcounter{table}{0}
\setcounter{section}{0}
\renewcommand{\theequation}{S\arabic{equation}}
\renewcommand{\thefigure}{S\arabic{figure}}
\section{Implementation details}
\begin{figure*}[ht]
    \centering
    \includegraphics[width=0.8 \linewidth]{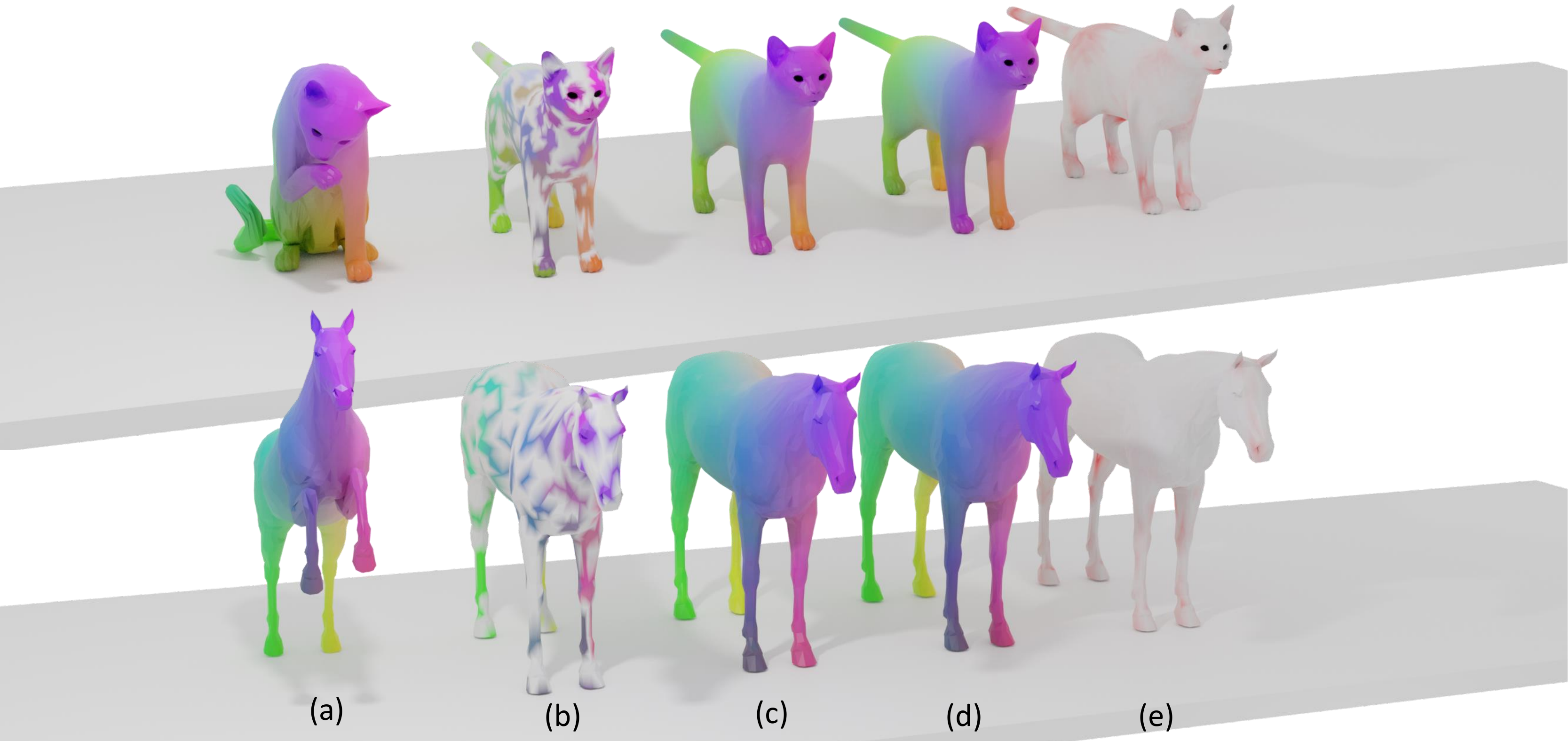}
    \caption{Results trained on SURREAL~\cite{surreal} and tested on Tosca~\cite{tosca}. In this experiment, we show the generalizability of the pipeline when trained on human shapes and tested on animal shapes. a) target shape. b) coarse matches c) source shape with the predicted correspondences d) ground-truth correspondences, e) normalized L2 error map.}
    \label{fig:supp03}
\end{figure*}
Here are further details which were not included in the experiment section of the paper.
\begin{itemize}
    \item  $d_{cut}$: 7
    \item Feature size of local graph: 64
    \item Feature size of shape graph: 64
    % \item Num. of GOT layers: 1
    \item Num. of Gated Feature Propagation (GFP): 2
    \item Weights first 30 epochs: $\gamma_{D}=1$, $\gamma_{M}=0$, $\gamma_{R}=1$
    \item Weights after 30 epochs: $\gamma_{D}=0.1$, $\gamma_{M}=1$, $\gamma_{R}=1$
\end{itemize}

\paragraph{Coarse-to-Fine Dense Matching}
We implement a simple algorithm based on functional maps to populate our matching to all mesh vertices densely. We provide 200 coarse matches as corresponding landmarks and create Wave Kernel Signature \cite{aubry2011WKS} of size 35 to find fine correspondences on all 6890 vertices.

\section{Domain Transfer}

% Animal to Human: train on TOSCA test on FAUST & SURREAL
To demonstrate the generalization ability of our network, we used the same model trained on SURREAL~\cite{surreal} dataset in section~4.3 to test on the TOSCA~\cite{tosca} horse class. The results are shown on Figure~\ref{fig:supp03} and table~\ref{tab:surreal_tosca}. The results suggest the strong domain transfer capability of our learned model to unseen shapes. 

\begin{table}[ht]
    \centering
    \begin{tabular}{c|c|c}
    \toprule
         & coarse error ($\downharpoonright$) & fine error ($\downharpoonright$) \\
         \midrule
        horse &  0.1988 & 0.0146 \\
        cat   &  0.2141 & 0.0223 \\
        \bottomrule
    \end{tabular}
    \caption{To evaluate the domain adaptability to new shapes, we use the model trained with SURREAL dataset to test on TOSCA dataset. We achieve equivalent results compared to the model trained on SMAL dataset with animal shapes.}
    \label{tab:surreal_tosca}
\end{table}

% \subsection{The effectiveness of GOT}
% The number of GRU
% The number of GOT
% The number of Sinkhorn iterations ?! yes
% Show both qualitative and quantitative results with & without regularization
% SFGOT1GRU2NFTPF Train: Surrfel Test:Fause GOT:1 GFP:2 No fine-tuning on point feature.
% Train 50 epochs
\section{Number of GOT and GFP}
\begin{figure*}[ht]
    \centering
    \includegraphics[width=0.8 \linewidth]{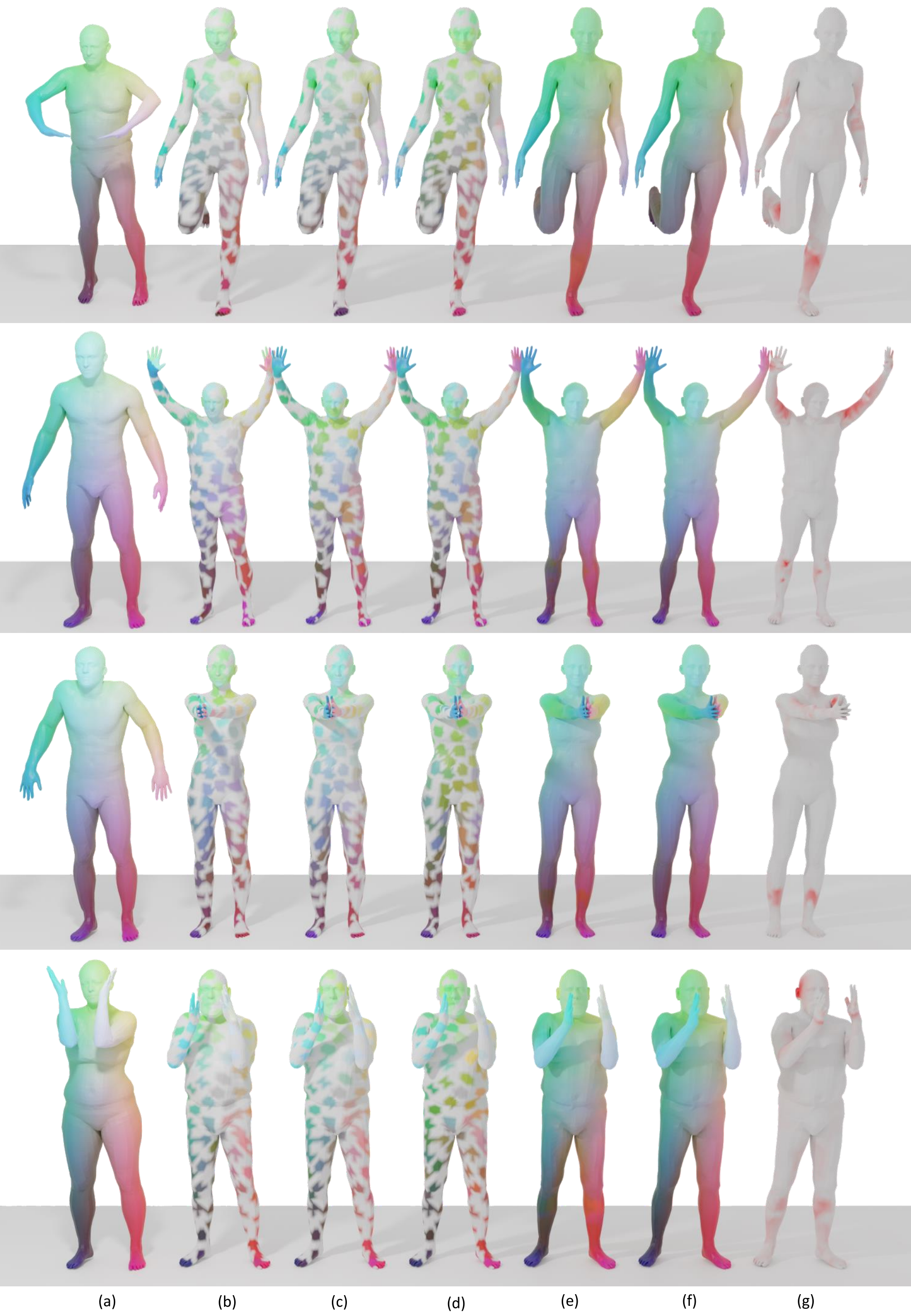}
    \caption{
    We show the visual difference of the ablation study on the number of GFP layers (corresponding to \#1~\#2 and \#3 on table~\ref{tab:abla_gotgru}. The human models from left to right are (a) The target shape map (b) The coarse matching result of model \#1, (c) the coarse matching result of model \#2, (d) the coarse matching result of model \#3, (e) the dense matching result of model \#2, (f) the ground truth dense matching result and (g) normalized L2 error map of model \#2.}
    \label{fig:supp01}
\end{figure*}

The proposed GOT consists of a Sinkhorn optimal transport layer followed by a GFP (See section~3.3). This involves two hyper-parameters, the number of message passing layers and the number of the total GOT operation. We ablate the effect on these two factors following the same setup of the ablation studies we provided in the main paper (Sec.~4.5). The results are shown on table~\ref{tab:abla_gotgru} and figure~\ref{fig:supp01}

By comparing \#2~\#4, it can be seen that the number of GFP does not linearly influence the result. Without GFP (\#1), the patch features are unaware of the local manifold, thus only focusing on matching similar features, resulting in a reasonable bijection rate and high error. With adequate GFP to enforce the regularity of the adjacent features, the network can achieve the best bijection rate and error. However, when the propagation is performed in several hops, the patch features are bound too widely to the local geometry and cannot provide a correct and distinctive matching. The same phenomenon is observed in the number of GOT operations. By comparing \#2, \#5, and \#6, we can observe that the more GOT operations, the lower the system's performance becomes. 

From \#3 and \#5, it can be seen that using two GOT operations with 2 GFPs is more effective than having one GOT with 4 GFPs. Note that these two settings are not identical. The former uses different confidence values on each GOT operation, while the latter uses the same confidence value to do 4 GFPs. Re-estimating the confidence values through Sinkhorn allows more flexible feature propagation. 

\begin{table}[ht]
    \centering
    \begin{tabular}{c|c|c|c|c}
    \toprule
        & N. GOT & N. GFP  & bij. rate ($\upharpoonright$) & err. ($\downharpoonright$)\\
    \midrule
        \#1 & 0 & 0 & 52.75 & 14.11 \\% & 23.74 & 2.441 \\ %SFGOT0NoBin  acc_coarse: 0.23736, acc_fine: 0.02441, bij_rate: 0.52750, geod_avg: 14.10975
        \#2 & 1 & 2 & \textbf{64.70} & \textbf{8.63} \\% & \textbf{13.26} & \textbf{2.413} \\ % SFGOT1GRU2NoBin acc_coarse: 0.13262, acc_fine: 0.02413, bij_rate: 0.64700, geod_avg: 8.63300,
        \#3 & 1 & 4 & 39.78 & 10.50 \\% & 15.30 & 2.452\\ %SFGOT1GRU4NoBin acc_coarse: 0.15300, acc_fine: 0.02452, bij_rate: 0.39775, geod_avg: 10.50000
        \#4 & 1 & 8 & 36.78 & 10.99 \\% & 15.92 & 2.451 \\ %SFGOT1GRU8NoBin  acc_coarse: 0.15916, acc_fine: 0.02451, bij_rate: 0.36775, geod_avg: 10.98800
        
        % 1 & 2 & \textbf{64.70} & \textbf{8.63} \\ %& \textbf{13.26} & \textbf{2.413} \\ % SFGOT1GRU2NoBin acc_coarse: 0.13262, acc_fine: 0.02413, bij_rate: 0.64700, geod_avg: 8.63300,
        \#5 & 2 & 2 & 45.78 & 9.59 \\% & 14.42 & 2.435 \\ % SFGOT2GRU2  acc_coarse: 0.14417, acc_fine: 0.02435, bij_rate: 0.45775, geod_avg: 9.59325,
        \#6 & 3 & 2 & 36.60 & 11.31 \\% & 16.69 & 2.426\\ % SFGOT3GRU2 acc_coarse: 0.16691, acc_fine: 0.02426, bij_rate: 0.36600, geod_avg: 11.30700
    \bottomrule
    \end{tabular}
    \caption{The ablation on the number of GOT and the number of GFP layers. As proven by bijectivity and error, the study \#2 with a single GOT and two layers of GFP performs the best.}
    \label{tab:abla_gotgru}
\end{table}

% \section{The Study on Qualitative Results}
% \ref{fig:supp01}

\section{Effect of GFP}
\begin{figure*}
    \centering
    \includegraphics[width=0.8\linewidth]{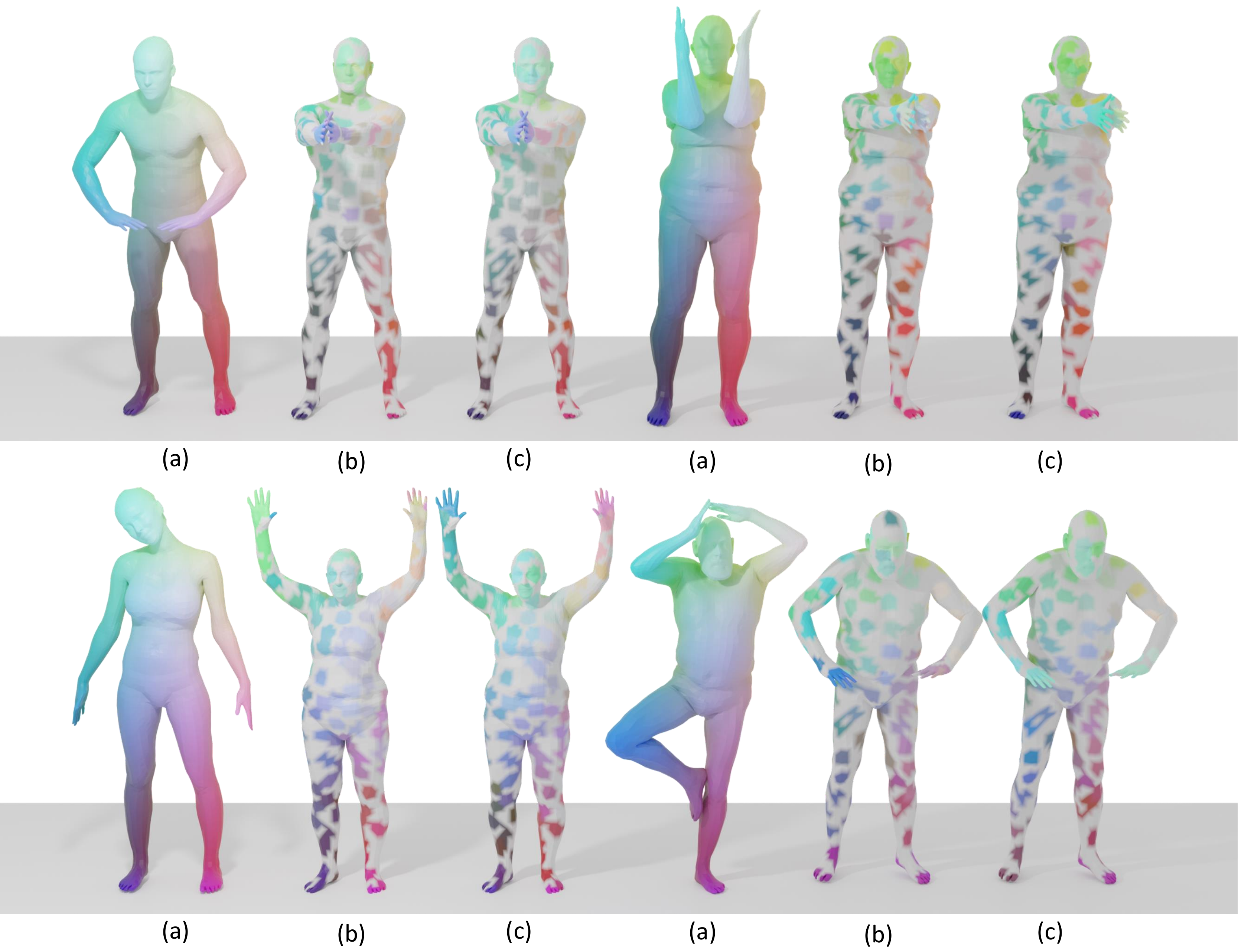}
    \caption{To show the effectiveness of our GFP module, we visualize correspondences before and after the GFP. Here a model is trained on SURREAL~\cite{surreal} and during evaluation on FAUST~\cite{faust}, we visualize the results from the first and final Sinkhorn. a) Correspondence map of target shape. b) Predicted correspondences from the first Sinkhorn (before GFP). c) Predicted correspondences from the final Sinkhorn (after the GFP). }
    \label{fig:supp02}
\end{figure*}

To study the effect of the GFP module on matching, we do an ablation on matching results before and after GFP. Here we train our network with one GOT and 2 GFPs and show the matching results of the first Sinkhorn versus the final Sinkhorn. Figure \ref{fig:supp02} shows the results of coarse matching on the Faust dataset. In Table \ref{tab:beforeafterGFP} we also compare coarse and fine geodesic errors with the outputs of each Sinkhorn layer. As visualized, the matching results after the GFP has lower error and less outliars. 
\begin{table}[ht]
    \centering
    \begin{tabular}{c|c|c}
    \toprule
         & coarse error ($\downharpoonright$) & fine error ($\downharpoonright$) \\
         \midrule
        Before GFP & 0.1315 & 0.0229 \\
        After GFP & 0.2896 & 0.0253 \\
        \bottomrule
    \end{tabular}
    \caption{We show the effect of GFP during evaluation on FAUST by comparing the matching error before and after the GFP module. Results are calculated from the output of the Sinkhorn.}
    \label{tab:beforeafterGFP}
\end{table}

% \input{supplementary}

%%%%%%%%% REFERENCES
% \newpage
% {\small
% \bibliographystyle{ieee_fullname}
% \bibliography{ms}
% }

\end{document}